\pdfoutput=1
\documentclass[11pt]{article}

\usepackage{acl}

\usepackage{times}
\usepackage{latexsym}
\usepackage{graphicx} 
\usepackage{booktabs}
\usepackage{multirow} 
\usepackage{tabularx}
\usepackage{enumitem} 
\usepackage{colortbl}
\usepackage{xcolor}

\usepackage[T1]{fontenc}
\usepackage[utf8]{inputenc}

\usepackage{microtype}
\title{When Raw Data Prevails: Are Large Language Model Embeddings Effective in Numerical Data Representation \\
for Medical Machine Learning Applications?} 

\author{
  \textbf{Yanjun Gao\textsuperscript{1,2}},
  \textbf{Skatje Myers\textsuperscript{2}},
  \textbf{Shan Chen\textsuperscript{3,5}},
  \textbf{Dmitriy Dligach\textsuperscript{4}},
\\
  \textbf{Timothy A Miller\textsuperscript{5}},
  \textbf{Danielle Bitterman\textsuperscript{3,6}},
  \textbf{Matthew Churpek\textsuperscript{2}},
  \textbf{Majid Afshar\textsuperscript{2}}
\\
\textsuperscript{1}University of Colorado, 
  \textsuperscript{2}University of Wisconsin Madison, \\
  \textsuperscript{3}Mass General Brigham, Harvard Medical School, 
  \textsuperscript{4}Loyola University Chicago, \\
  \textsuperscript{5}Boston Children Hospital, Harvard Medical School, \\
  \textsuperscript{6}Dana-Farber Cancer Institute 
\\
  \small{
    \textbf{Correspondence:} \href{mailto:ygao@medicine.wisc.edu}{ygao@medicine.wisc.edu}
  }
}

\begin{document}
\maketitle
\begin{abstract}
The introduction of Large Language Models (LLMs) has advanced data representation and analysis, bringing significant progress in their use for medical questions and answering. Despite these advancements, integrating tabular data, especially numerical data pivotal in clinical contexts, into LLM paradigms has not been thoroughly explored.
In this study, we examine the effectiveness of vector representations from last hidden states of LLMs for medical diagnostics and prognostics using electronic health record (EHR) data. We compare the performance of these embeddings with that of raw numerical EHR data when used as feature inputs to traditional machine learning (ML) algorithms that excel at tabular data learning, such as eXtreme Gradient Boosting. We focus on instruction-tuned LLMs in a zero-shot setting to represent abnormal physiological data and evaluating their utilities as feature extractors to enhance ML classifiers for predicting diagnoses, length of stay, and mortality. Furthermore, we examine prompt engineering techniques on zero-shot and few-shot LLM embeddings to measure their impact comprehensively. Although findings suggest the raw data features still prevails in medical ML tasks, zero-shot LLM embeddings demonstrate competitive results, suggesting a promising avenue for future research in medical applications. 

\end{abstract}

\section{Introduction}

\begin{figure}
    \centering
    \includegraphics[scale=0.46]{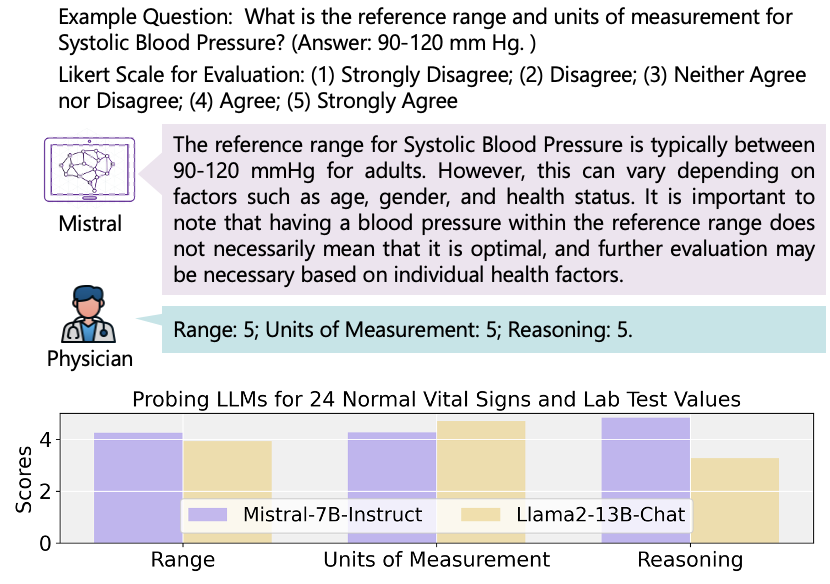}
    \vspace{-.1in}
    \caption{\small Physician Evaluation of LLMs' Knowledge on Normal Vital Sign and Lab Test Values. This experiment probes Mistral-7B-Instruct and Llama2-13B-Chat on reference ranges for 24 vital signs and lab tests. Results show these models have a strong understanding of normal medical values, crucial for clinical applications. Table~\ref{tab:feat_and_templates} listed all 24 feature names, and more output examples are in Appendix~\ref{sec:prob_examples}. }
    \label{fig:probing}
\end{figure}
Numerical data plays a pivotal role across various domains. For instance, much of the data used for analytics from electronic health records (EHRs) are numerical values in tabular formats, documenting patient demographics (e.g., age), vital signs, laboratory tests, and nurse assessments. Utilizing numerical data for predictive modeling has been instrumental in facilitating accurate diagnoses~\cite{pang2021cehr}, risk stratifying~\cite{zeiberg2019machine,green2018comparison}, and outcome predictions~\cite{akel2021less,chang2019machine} in healthcare. Machine learning (ML) classifiers like gradient boosted~\cite{chen2016xgboost} have excelled in these tasks for making accurate clinical predictions~\cite{RN44,lolak2023comparing,moore2022xgboost}. 

\begin{figure*}
    \centering
    \includegraphics[scale=0.38]{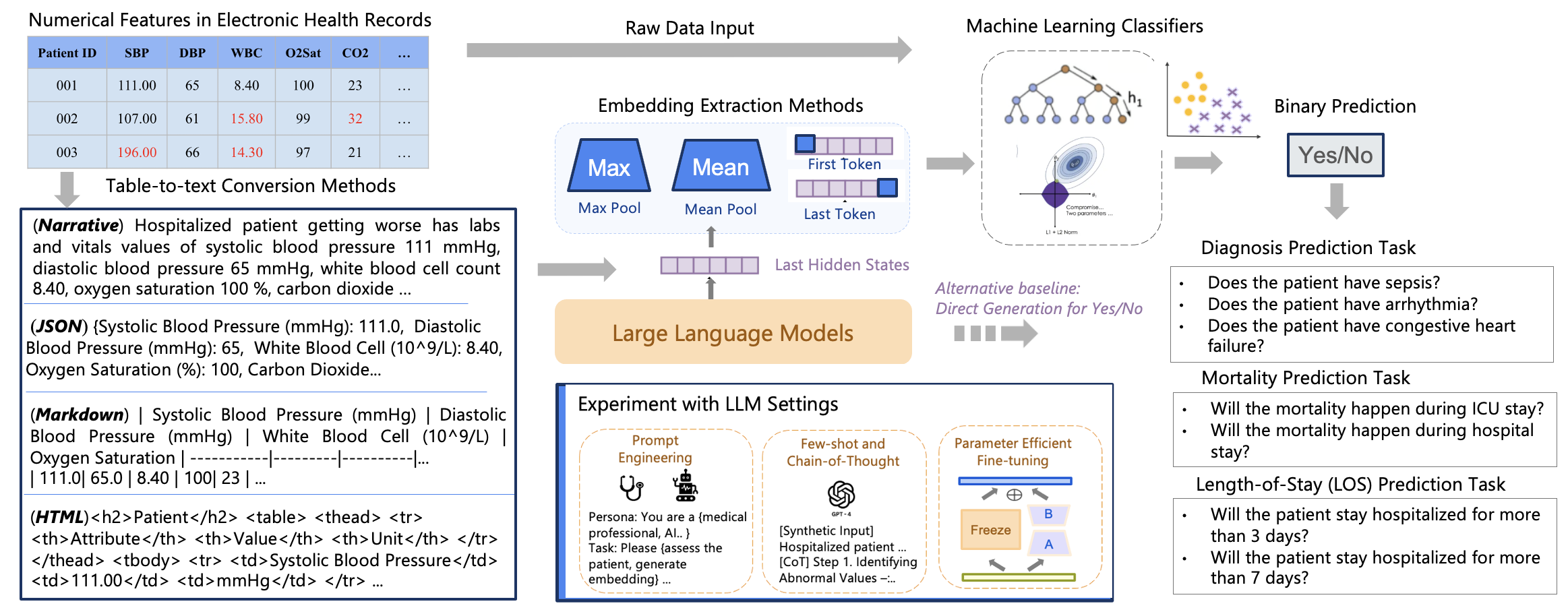}
    \vspace{-.13in}
    \caption{\small This study investigates the feasibility of using LLM embeddings for numerical EHR data features representation in medical machine learning applications.  To use LLMs, raw features are transformed into queries via templates. Under a zero-shot setting, these queries are encoded into embeddings for ML classification. We explore the effects of prompt engineering, few-shot learning using synthetic data generation, and parameter efficient tuning on LLM embeddings.}
    \label{fig:study}
\end{figure*}

%The advent of Large Language Models (LLMs) has transformed data representation. 
Recent work shows Large Language Models (LLMs)' vast potential on text generation over structured data input, including Chain-of-Thought (CoT) reasoning over tabular data~\cite{zheng-etal-2023-chain}, classification on diseases~\cite{hegselmann2023tabllm}. LLMs have also exhibited exceptional promise in medical NLP tasks, evident in their stellar performance in the United States Medical Licensing Examination (MedQA)~\cite{nori2023can}. 
However, the use of embedding representations, particularly for medical diagnostics and outcome predictions using standard EHR numerical data, remains largely unexplored. In these areas, raw data inputs have traditionally dominated feature representation for ML algorithms before the era of LLMs. This is exemplified by their use in critical applications such as mortality prediction and early sepsis warnings~\cite{deng2022evaluating,hou2020predicting}, and patient infection~\cite{bashiri2022identifying,bhavani2020development}. The potential of LLM-derived features as a viable alternative to raw data features in ML applications is still unclear.  
%It remains unclear whether features derived from LLMs can serve as a viable alternative to raw data features for ML applications, considering their extensive training and proven success in medical question-answering.

This study aims to address this knowledge gap by examining the use of LLM embeddings for EHR numerical data representation in ML algorithms. Although LLMs are renowned for text generation, their embeddings may offer multiple advantages, such as leveraging LLMs' pre-trained knowledge and sophisticated text understanding to enhance domain-specific tasks. Moreover, using LLMs to represent tabular data allows for a unified model that encodes both structured and unstructured text in EHRs, seamlessly integrating and contextualizing information across modalities, such as embedded tables in clinical notes~\cite{soenksen2022integrated,kline2022multimodal}. 

Our work presents novel examination of the impact of different formats and embedding methods on LLM last layers and ML classifiers. We focus on open-source, zero-shot LLMs suitable for single-GPU systems, considering the resource limitations prevalent in many hospitals and academic research settings. To establish a foundation for this work, we probed Mistral-7B-Instruct and Llama2-13B, two open-source, general-domain LLMs, for their knowledge of reference ranges for vital signs and lab test values. We directly asked about the standard physiological values and units of measurement for 24 EHR features identified as critical predictor variables for detecting clinical deterioration~\cite{akel2021less}. As in Figure~\ref{fig:probing}, physician judgment indicates that LLMs possess this knowledge, providing initial evidence for further investigation. %, as shown in Figure~\ref{fig:probing}.

Our study utilizes three clinical prediction tasks derived from two independent EHRs and four ML classifier input settings. We investigate the impact of table-to-text conversion formats, embedding extraction methods, prompt engineering, and few-shot techniques, along with early results from parameter-efficient fine-tuning, on the quality of LLM embeddings. Our main contributions are threefold:
\vspace{-\topsep}
\begin{itemize}[leftmargin=*,noitemsep]
    \setlength{\itemsep}{0pt}
    \item We present a comprehensive study exploring various factors that influence the performance of numerical EHR feature embeddings generated by LLMs for medical ML applications.
    \item Our findings show that while LLM embeddings paired with XGB classifiers can achieve performance comparable to traditional raw data features on some tasks, performance gaps persist, necessitating further improvements to maximize their effectiveness.   
    \item We discuss the efficiency and robustness of LLM feature representation for numerical data versus raw data in training ML classifiers. 
\end{itemize}
\vspace{-\topsep}
Results show that, despite external evidence indicating that LLMs possess extensive knowledge of medical facts, extracting usable representations of this knowledge for downstream tasks will require significant additional methodological progress. %Nonetheless, LLM embeddings paired with XGB classifiers achieve performance comparable to raw data features and demonstrate greater robustness than direct LLM generative approaches. 

\section{Related Work}

%\paragraph{LLM for numerical data representation} 

Recent studies highlight LLMs in tabular data analysis: \citeauthor{hegselmann2023tabllm} (\citeyear{hegselmann2023tabllm}) introduces TableLLM, which converts tables to text using a manual template. \citeauthor{zheng-etal-2023-chain} (\citeyear{zheng-etal-2023-chain}) studies CoT reasoning over tables. \citeauthor{akhtar2023exploring} (\citeyear{akhtar2023exploring}) examines the abilities of LLMs on numerical data understanding. \citeauthor{zhu2024prompting} (\citeyear{zhu2024prompting}), closest to our work, explores zero-shot LLM for structured longitudinal EHR data and finds that GPT-4 can outperform XGB on clinical prediction tasks. 
Our study, however, uniquely focuses on open-box LLM embeddings for enhancing ML algorithms. %Our study, however, has a unique focus on LLM embeddings for ML algorithms. 

Raw EHR data are commonly used in medical ML applications, as found by a survey on medical ML research~\cite{si2021deep}. They noted that labs and vital signs as frequent data types for patient representation learning. \citeauthor{RN44} (\citeyear{RN44}) introduces an XGB algorithm predicting clinical deteriorations using EHR features like demographics and lab values. \citeauthor{wang2020mimic} (\citeyear{wang2020mimic}) used 104 clinical EHR features across various ML algorithms to establish baselines for clinical tasks such as mortality predictions. Our work uses the same dataset and tasks as \cite{wang2020mimic} to compare LLM embeddings against traditional ML classifier outcomes on the same raw data feature baseline.

%\paragraph{Machine learning using raw data features from EHR} \textcolor{red}{To-be-added}
%concentrate on the diagnosis prediction task, examining the potential correlation between LLM-generated scores and those from XGBoost. %thereby navigating a novel intersection between tabular data understanding and predictive analytics in healthcare. 
\section{Datasets and Tasks} 

\subsection{Diagnosis prediction for clinical deterioration}
Early warning systems often use rule-based and ML algorithms to identify patients at risk of deterioration or death without providing diagnoses~\cite{churpek2014multicenter,kipnis2016development}. To address this, experts from multiple hospitals created a dataset that labels the diagnoses for patients who had a clinical deterioration event during their hospitalization. These expert-annotated diagnoses were performed with a full review of the EHR and served as the labels for our training data. Twenty-four tabular data features including demographics, vital signs, labs, interventions, and nursing assessments were extracted from the structured EHR (eg. tabular data). They were previously identified as critical variables for clinical deterioration~\cite{akel2021less}. The final datasets encompassed EHR data from 660 adult patients in medical-surgical ward within a U.S. health system. The primary diagnoses were Sepsis, Arrhythmia (Arrhy.), and Congestive Heart Failure (CHF) volume overload, with prevalence rates of 43.18\% for Sepsis, 15.30\% for Arrhythmia, and 11.82\% for CHF, respectively. We used 5-fold validation on all 660 samples to generate five distinct test sets.  %The primary diagnoses contributing to clinical deterioration in this cohort are Sepsis, Arrhythmia (Arrhy.), and Congestive Heart Failure (CHF) volume overload. The prevalence rates within this cohort for these conditions are 43.18\% (285 of 660) for Sepsis, 15.30\% (101 of 660) for Arrhythmia, and 11.82\% (78 of 660) for CHF, respectively. Details about the demographic characteristics of the patients, including gender, age, and race, are included in Appendix. 
\footnote{The dataset used in this study has been detailed in ~\cite{churpek2024causes}. Details about the demographic characteristics of the patients, including gender, age, and race, are included in Appendix.  } 
%We implemented 5-fold validation on the entire dataset of 660 samples to create five distinct test sets.  

\begin{table}
\scriptsize 
\begin{tabularx}{\columnwidth}{X}  % One column to auto-wrap, one fixed width
\toprule

\textbf{Input features} Age, Systolic Blood Pressure, Diastolic Blood Pressure, Oxygen Saturation, Temperature in Celsius, Proton Pump Inhibitor, Alert, Voice, Pain, Unresponsive Scale (AVPU), Albumin, Alkaline Phosphatase, Anion Gap, Total Bilirubin, Blood Urea Nitrogen, Blood Urea Nitrogen to Creatinine Ratio, Calcium, Chloride, Carbon Dioxide, Creatinine , Serum Glucose, Hemoglobin, Platelet Count, Potassium, Serum Glutamic-Oxaloacetic Transaminase, Sodium, Total Protein, White Blood Cell Count \\
\textbf{Target prediction} Sepsis, Arrhythmia, Congestive Heart Failure (CHF) Volume Overload \\ 
\bottomrule 
\end{tabularx}
    \vspace{-.12in}
\caption{\small Raw clinical data features from the EHR for diagnosis prediction task.  }
\label{tab:feat_and_templates}
\end{table}

%\paragraph{Data serialization} 
Table~\ref{tab:feat_and_templates} outlines the structured input features from the cohort EHR dataset and target diagnoses utilized in our analysis. The input features comprised a comprehensive set of clinical data points including demographic information like age, vital signs such as Systolic and Diastolic Blood Pressure, and body Temperature, as well as a range of serum laboratory tests including electrolytes, liver function panel, renal function, red blood counts, etc. These inputs served as predictors and are relevant findings in making diagnoses like Sepsis, Arrhythmia, and CHF. Despite its smaller sample size, this EHR dataset includes physicians' manual chart reviews and carefully curated data, providing accurate annotations for patient diagnoses.  %Even though it was smaller in sample size, it is an EHR dataset with physicians' manual chart reviews and carefully curated data, representing accurate annotation on the diagnosis of patients. 

%Additionally, the table includes a data serialization template used to format these varied clinical measurements into a standardized query. This template details the format in which data from the EHR dataset are presented, integrating both laboratory results and vital signs into a single descriptive snapshot of a patient's current state. Each placeholder in the template is populated with actual data points from patient records, facilitating precise and contextualized diagnostic predictions. Using this template, we can transform tabular EHR data into a format suitable for input into an LLM, from which we then generate embeddings. 

\subsection{Mortality and length-of-stay prediction}

%MIMIC dataset; 104 lab test and vital signs attributes, 

The MIMIC-III dataset, derived from the EHR of the Critical Care Units (ICU) at Beth Israel Deaconess Medical Center, has been utilized extensively in research~\cite{johnson2016mimic}. \citeauthor{wang2020mimic} (\citeyear{wang2020mimic}) further developed an open-source pipeline for extracting, preprocessing, and representing data from the MIMIC-III database, namely MIMIC-Extract. This pipeline aggregates various data types, such as tabular demographic data available at admission, vital signs with repeated measures, laboratory test results, time-varying intervention signals, and prediction labels needed for clinical tasks. MIMIC-Extract introduces two clinical prediction tasks: mortality and length-of-stay (LOS) predictions. The mortality prediction task uses tabular data from the first 24-hour window of a patient's ICU stay to predict mortality as a binary classification task. The LOS prediction task, in contrast, determines whether a patient's stay will exceed three (LOS 3) or seven days (LOS 7) based on the same 24-hour data period. Importantly, to avoid competing risk outcomes between death and LOS, patients who died within the 3- or 7-day LOS window were excluded from the LOS prediction. 

We adopted the same data partitioning used in~\citep{wang2020mimic}, comprising 16,700, 2,394, and 4,790 patient records for the training, development, and testing sets. Each patient record includes 104 time-varying tabular data features. More detailed demographic information can be found in the MIMIC-Extract study~\cite{wang2020mimic}. 
The labels in the MIMIC-Extract dataset are highly skewed, with positive label distributions of 42.82\% for LOS 3, 7.66\% for LOS 7, 10.27\% for Mort Hosp, and 7.10\% for Mort ICU.

\begin{table}
\scriptsize  
\begin{tabularx}{\columnwidth}{X}  % One column to auto-wrap, one fixed width
\toprule
\textbf{Diagnosis dataset} Hospitalized patient of age \textit{[value]} getting worse has labs and vitals values of systolic blood pressure \textit{[value]} mmHg, diastolic blood pressure \textit{[value]} mmHg, oxygen saturation\textit{[value]} \%, body temperature \textit{[value]} celsius degree, ...  total protein \textit{[value]}, white blood cell \textit{[value]}. What are the diagnoses for this patient?  \\
\midrule 
\textbf{MIMIC-Extract} Hospitalized patient with lab and vital signs available: in the past 24 hours, the observed alanine aminotransferas values are [\textit{list of unique values sorted by temporal order}], albumin values are [\textit{list of unique values sorted by temporal order}], anion gap values are [\textit{list of unique values sorted by temporal order}]...Predict if the patient mortality will occur in-hospital. \\
\bottomrule 
\end{tabularx}
\vspace{-.1in}
\caption{\small The template for \textsc{narrative} serialization method for diagnosis prediction dataset (top) and MIMIC-Extract dataset (bottom).  }
\label{tab:mimic_template}
\end{table}

\begin{table}[h]
\small 
    \centering
    \begin{tabular}{lll} \toprule 
        Dataset & Size & Average input tokens \\ \midrule 
        Diagnosis & 660 & 346.97 $\pm$ 2.21  \\
        MIMIC-Extract & 23,884 &1829.57 $\pm$ 497.02 \\ 
        \bottomrule 
    \end{tabular}
    \vspace{-.1in}
    \caption{\small Dataset description}
    \label{tab:my_label}
\end{table}
%We are testing LLM embeddings with machine learning classifiers  

%\subsection{Evaluation setup}
 
\section{Methods and Experiment Setup}

%Our investigation focuses on two types of input data for medical ML applications: raw data features and LLM embeddings. The raw data features are structured into Pandas DataFrames for efficient tabular data handling and preprocessing. This format integrates easily with ML pipelines, directly using numerical scores and variable names for training.
Figure~\ref{fig:study} illustrates the study overview and experiment setup. We began with a patient's tabular data input, represented using the Pandas DataFrame data structure (\textit{raw data}). This raw data was converted to text using four distinct conversion methods, detailed in \S\ref{sec:conversion}, and LLM encoded the converted text, with the last hidden states extracted to generate embedding features (\S \ref{sec:emb_ext}). These embeddings were subsequently used to train various ML classifiers on two datasets for binary prediction tasks.

We started with zero-shot, off-the-shelf LLMs for experiments (\S\ref{sec:llm_selection}). We then investigated the impact of prompt-engineering techniques and few-shot learning configurations on the embeddings and subsequent predictions (\S \ref{sec:prompt}). An initial investigation was also conducted to assess the effects of parameter-efficient fine-tuning on LLM embeddings for ML tasks, focusing on two of the models (\S\ref{sec:qlora}).

As baselines, we included traditional ML classifiers trained directly on raw tabular data inputs. To benchmark the effectiveness of LLM embeddings, we used randomly initialized embeddings of the same size as the LLM-generated embeddings.  

\subsection{Table-to-text conversion} 
\label{sec:conversion}
We employed four different methods to convert EHR tables into input formats for LLMs: \textsc{Narratives, Json, html}, and \textsc{Markdown}. \textsc{Narratives} provide a continuous text description of patient data, offering context and readability similar to clinical notes~\cite{yu2023unified}. \textsc{Json} structures the data hierarchically, making it easy to parse and interpret programmatically~\cite{zhao2023large}. \textsc{Html} format leverages web-based structures to present the data with tags~\cite{garcia2024medmt5}. \textsc{Markdown} offers a lightweight markup language that provides formatting while remaining readable in plain text~\cite{zhao2023large}.

Table~\ref{tab:mimic_template} includes two \textsc{narratives} templates used to format these varied clinical measurements into a standardized query. These templates detail the format in which data from the EHR dataset are presented, integrating both laboratory results and vital signs into a single descriptive snapshot of a patient's current state. Each placeholder in the template is populated with actual data points from patient records, facilitating the transformation of tabular EHR data into a format suitable for LLM input, from which we then generate embeddings. 

The primary distinction between the templates for the diagnosis prediction dataset and the MIMIC-Extract dataset lies in the types of values incorporated. For diagnosis prediction, data are values collected immediately before the early warning system triggers for clinical deterioration. In contrast, MIMIC-Extract tasks include laboratory and vital signs data from the 24 hours prior to the event. We extracted all unique values observed during the first 24 hours of ICU admission in chronological order, compiling these into a list format. If a feature has no observations, it is omitted, resulting in variable length sequences. %This approach enables the LLM to handle diverse clinical data effectively. 

\subsection{Embedding extraction methods}
\label{sec:emb_ext}
This section introduces the methods used to convert input text to fixed-size vector for ML input. We focused on the \textit{last hidden states} of LLMs (as in~\cite{lu2021parameter}), and employed three different embedding extraction methods: 
\textbf{Max Pooling} captures the most salient features by taking the maximum value across all token embeddings for each dimension~\cite{bao2023docasref};
\textbf{Mean Pooling} computes the average value of the token embeddings, providing a balanced representation reflecting the overall content~\cite{ram2023you}; \textbf{Last Token} uses the embedding of the last token as the representation, capturing the concluding context or final summary~\cite{shani2023towards,fu2023estimating}. We included embeddings extracted from \textbf{first token} as a reference point despite it is not ideal due to the nature of decoder-only models. 

%\subsection{Overview of machine learning classifiers}
%\label{sec:ml_cls}
Our choice of ML classifiers comprised two tree-based methods and a linear model to provide a comprehensive assessment of various predictive approaches. Specifically, we utilized eXtreme Gradient Boosting (XGB)\cite{chen2016xgboost} and Random Forest (RF)\cite{breiman2001random} as our tree-based classifiers due to their robustness and efficiency in handling diverse datasets with accuracy. Additionally, Logistic Regression with regularization (LR) as our linear model was chosen for its effectiveness in preventing overfitting via Ridge and Least Absolute Shrinkage and Selection Operator regularization\cite{zou2005regularization}. Together, these classifiers form a balanced baseline setup that caters to both non-linear and linear decision boundaries in our data.

\subsection{Selection of LLMs}
\label{sec:llm_selection}
We assessed a mix of general-domain models and models trained on medical text. Three widely-used, general-domain LLMs that have been instruction-finetuned are Mistral-7B-Instruct-v0.1~\cite{jiang2023mistral}, Llama2-13B-chat-hf, Llama2-70B-chat-hf~\cite{touvron2023llama}, and Llama3-8B-instruct~\cite{llama3_github}. These models are compatible with one Nvidia 80GB A100 GPU, making them popular choices among available LLMs. For the domain specific LLM, We selected Meditron-7B~\cite{chen2023meditron}, a Llama2-7B based model continuously pretrained on medical text. We also included ClinicalBERT~\cite{alsentzer2019publicly}, pre-trained on MIMIC EHR text, representing encoders pre-trained on clinical text baseline compared to decoder-only LLMs.

\subsection{Prompt design and few-shot learning} 
\label{sec:prompt}

Because the majority of LLMs we tested are instruction-tuned and require varying input formats, we utilized the chat templates to ensure proper integration of input data~\cite{zheng-2024-chat-templates}. 
In our study, the default setting involves including only the task-relevant question (shown on the right side of Fig~\ref{fig:study}) in the system message and the converted EHR data in the user input, without additional system instructions, predefined personas, or other context. Given that instruction-tuned LLMs are known to be sensitive to system instructions, we designed four system instructions that vary by persona {\small (medical professional, AI system)}, tasks{\small ~(assess patients, generate embeddings for ML classifiers)}, thinking style{\small ~(chain-of-thoughts)}, and question type{\small ~(general assessment, binary question)}, enabling us to explore the influence of prompt characteristics on the embeddings. All prompts were paraphrased for better perplexity scores, following prompt optimization strategies~\cite{gonen2023demystifying,lu2023makes}.  

Two few-shot settings were explored besides zero-shot prompt engineering. We generated synthetic data for diagnosis prediction, by prompting GPT-4 to generate values based on the attribute names in Table~\ref{tab:feat_and_templates}. For each target diagnosis, GPT-4 generated one example confirming the diagnosis (positive) and one example negating it (negative). Moreover, GPT-4 was asked to generate CoT explanations identifying abnormal values and their clinical significance. An expert physician and clinical informaticist reviewed these synthetic data pairs for quality assurance. The complete set of prompts are presented in Table~\ref{tab:prompts}.   
%Our prompt engineering strategy followed previous work demonstrating the use of perplexity to select optimal prompts~\cite{gonen2023demystifying}. We designed a set of meta-language prompts with different personas and task scenarios, presented in Appendix. For each meta-prompt, five paraphrases was generated using GPT-4. The perplexity score for each paraphrase was computed to gauge naturalness and fluency, then the one with the lowest perplexity was selected as the final prompt. 

%We also investigated whether few-shot learning can enhance LLM embeddings, using synthetic examples generated from GPT-4. We focus on the diagnosis prediction task, which involves relatively short inputs with 24 clinical features, as opposed to the more complex MIMIC-Extract dataset, which includes 104 features and 24 data points per feature, making it more challenging to verify the quality of synthetic data generation. 

%For the synthetic data examples, we instructed GPT-4 to generate values based on the attribute names in Table~\ref{tab:feat_and_templates}. For each target diagnosis, GPT-4 generated one example confirming the diagnosis (positive) and one example negating it (negative). Additionally, GPT-4 was asked to generate CoT explanations identifying abnormal values and their clinical significance. An expert physician and clinical informaticist reviewed these synthetic data pairs for quality assurance. The few-shot examples, along with the previously designed prompts, were prepended to each actual input sample, and we tested the LLM on two few-shot settings, with and without CoT. 

\subsection{Parameter efficient fine-tuning} 
\label{sec:qlora}
While our paper primarily focuses on evaluating zero-shot LLMs for numerical feature representation, we included parameter-efficient fine-tuning experiment to suggest future directions for improvement. We employed QLoRA~\cite{dettmers2024qlora} on Mistral-7B-Instruct and Llama3-8B-Instruct, using the MIMIC-Extract dataset due to its larger training set compared to the diagnosis dataset. We trained Mistral with a sequence classification head on top, saving checkpoints with the lowest validation loss. Based on validation performance, we optimized the (q, k, v, o) layers with $r=16$, a learning rate of 3e-5, and a LoRA dropout of 0.1. Each model was trained for 3 epochs with early stopping to prevent overfitting. %Training on MIMIC data took an average of 4.5 hours per model 

\begin{table*}[h]
\centering
\scriptsize 
\begin{tabular}{lllll}
\toprule
\textbf{Model}  & \textbf{Sepsis AUROC}  & \textbf{Arrhythmia AUROC}  & \textbf{CHF AUROC} & Average \\ 
& (95\% CI) & (95\% CI) & (95\% CI)& (95\% CI) \\
\midrule
%\multicolumn{5}{c}{LLM Generation Baseline} \\ \midrule 
\multicolumn{5}{c}{Raw Data Features Baseline} \\ \midrule 
LogisticRegression &  71.10 (67.01, 75.18) & \textbf{74.40} \textbf{(69.35, 79.56)} & 54.79 (47.74, 61.79) & 66.76 (61.37, 72.18) \\
RandomForest &  65.26 (61.79, 68.48) & 53.07 (50.58, 55.80) & 50.89 (49.01, 53.43) & 56.41 (53.79, 59.24) \\
%XGB & \textbf{71.17 (67.06, 75.11)} & \textbf{76.49 (71.32, 84.13)} & 58.47 (51.36, 65.15) & \textbf{68.71 (63.25, 74.80)} \\
%XGB$_{RandMat}$ &54.01 (49.89,58.44) & 49.65(44.02,54.62) & \\ 
 \midrule
\multicolumn{5}{c}{LLM embedding + XGB classifier} \\ \midrule 
Random & 54.01 (49.89,58.44) & 49.65(44.02,54.62) & 50.02 (47.13, 52.29)& 51.22 (47.01, 55.19) \\ 
Mistral-7b-Instruct$_{best}$ & \cellcolor{green!30}\textbf{71.12} \textbf{(67.54, 74.92)} & \cellcolor{green!30}68.00 (61.52, 73.93) & \cellcolor{green!30}51.80 (44.48, 58.65) & \cellcolor{green!30}63.40 (57.73, 68.77) \\
%Mistral$_{all}$ & 63.21 (61.68, 64.83) & 56.32 (51.58, 61.06) & 52.16 (48.11, 56.20) & 63.40, (57.73, 68.77) \\
Llama3-8b-Instruct$_{best}$ & \cellcolor{green!30}63.84 (57.31, 69.87) & \cellcolor{green!30}71.08 (65.69, 75.87) & \textbf{63.84 (56.77, 70.37)} & \cellcolor{green!30}66.25 (60.15,72.35)\\ 
Llama2-13b$_{best}$ & \cellcolor{green!30}66.02 (61.64, 70.32) & 58.62 (52.62, 64.46) & 49.69 (48.83, 62.58) & \cellcolor{green!30}58.11 (54.36, 65.79) \\
Llama2-70b-chat$_{best}$ & \cellcolor{green!30}68.57 (63.88, 71.53) & 69.15 (67.08, 71.17) & \cellcolor{green!30}53.87 (49.83, 58.52) & 63.86 (60.93, 67.07) \\
Meditron$_{best}$ & 66.74 (62.30, 66.15) & \cellcolor{green!30}72.26 (65.28, 77.43) & \cellcolor{green!30}58.11 (50.64, 64.48) & \cellcolor{green!30}63.90 (58.28, 65.45) \\
ClinicalBERT & 58.80 (54.44, 63.04) & 64.91 (61.84, 70.27) & \cellcolor{green!30}49.67 (41.94, 57.51) & 57.79 (52.74, 63.11) \\
 \midrule
\multicolumn{5}{c}{LLM embedding + Logistic Regression classifier} \\ \midrule 
Random & 49.58 (47.68, 51.12) & 49.22 (48.09, 50.43) & 49.36 (47.12 51.06) & 49.39 (47.63, 50.87) \\
Mistral-7b-Instruct$_{best}$ & 62.61 (58.17, 66.95) & \cellcolor{green!30}69.59 (64.67, 74.71) & 48.98 (42.96,55.62) & 60.39 \cellcolor{green!30}(55.27, 65.76) \\ 
Llama3-8b-Instruct$_{best}$ & \cellcolor{green!30}66.54 (62.32, 70.62) & \cellcolor{green!30}70.22 (64.82, 74.11) & \cellcolor{green!30}63.52 (55.91,69.20) &  \cellcolor{green!30}66.76 (61.50, 72.02)  \\ 
Llama2-13b-chat-hf$_{best}$ & \cellcolor{green!30}66.95 (62.82, 70.88) & 66.04 (60.04, 71.22) & \cellcolor{green!30}58.54 (52.09, 65.09) & \cellcolor{green!30}63.84 (58.32, 69.06) \\ 
Llama2-70b-chat-hf$_{best}$ & \cellcolor{green!30}69.50 (65.37, 73.43) & 68.11 (61.75, 70.57) & \cellcolor{green!30}62.72 (56.47, 68.39) & \cellcolor{green!30}66.78 (61.20, 70.80) \\ 
Meditron$_{best}$ &  \cellcolor{green!30}66.91 (62.83, 71.09) & \cellcolor{green!30}68.61 (63.49, 73.72) & \cellcolor{green!30}57.60 (51.02, 63.89) & \cellcolor{green!30}64.37 (59.11, 69.90) \\  
ClinicalBERT & 47.28 (43.07, 51.63) & 44.62 (38.79, 50.29) & 46.98 (42.96, 55.62) & 46.29 (41.61, 52.51) \\ 
\bottomrule
\end{tabular}
\vspace{-.1in}
\caption{\small Comparing raw data features and LLM embeddings features for ML classifiers on Diagnosis dataset. We report the best AUROC scores from LLM embedding across various embedding extraction and table-to-text conversion methods. The ``Random'' row indicates the randomly initialized embedding input. For ClinialBERT, we used [CLS] token embedding as the final representation. We use green color to highlight the LLM+ML results where it has CI overlapping with the best results (in bold fonted text). }
\label{tab:ml_baselines_sepsis_arrhythmia_chf_prediction}
\end{table*}

\subsection{Experiment setup}

%We implemented the grid-search to find the best set of hyperparameters.
We used a 5-fold cross-validation on the diagnosis dataset (660 patient records), resulting in 528 patients for training and 132 for testing per fold. For mortality and LOS prediction tasks from MIMIC-Extract data, we followed the data split from~\cite{wang2020mimic}. We evaluated performance using Area Under the Receiver Operating Characteristic (AUROC) with 95\% confidence intervals (CI).

For all ML classifiers, we determined the best parameters through grid search on the validation set. Specifically, we tuned the number of estimators, maximum depth, learning rate, and minimum child weight for XGB classifiers, and alpha and L1 ratio for LR classifiers (see Appendix~\ref{sec:param_grid}). For LLMs under 13B, the maximum input length was 1042 for the diagnosis dataset and 3076 for the MIMIC dataset, resulting in a 4096-dimensional embedding. For 70B LLM, the max input length was 1500 and 4-bit quantization was set to avoid GPU memory errors, producing an 8192-dimensional embedding. All experiments ran on an Ubuntu server with an Nvidia 80GB A100 GPU.     %To train ML classifiers, we use 5-fold cross validation and conduct parameter searching on the training set on Diagnosis dataset given the dataset only contains 660 samples. 

\section{Results}
%\input{latex/impact_format_embedding}
%\section{RQ1: Zero-shot LLM Embedding vs. Raw Data Features for ML classifiers}

\subsection{Main results for diagnosis prediction}
 \label{sec:rq1}

Table~\ref{tab:ml_baselines_sepsis_arrhythmia_chf_prediction} presents AUROC scores for predicting Sepsis, Arrhythmia, and CHF with different ML models, demonstrating the effects of using LLM embeddings compared to raw data features. Traditional ML classifiers with raw features demonstrate high AUROC scores. %XGB with raw data features stood out in the baseline, demonstrating the highest AUROC for Sepsis and Arrhythmia, and the highest average AUROC across all diagnoses. LR and RF, while using raw data input, showed moderately lower effectiveness.

For LLM embeddings with zero-shot setting, we observed performance gain over a randomly initialized embedding approach into XGB with substantial gains in all decoder LLMs. ClinicalBERT was the only model, as an older pre-trained encoder, that did not show notable performance gains over the randomly initialized embedding model. Further, Mistral embedding with XGB classifiers achieved a competitive AUROC of 71.12 (vs. 71.16 of raw data with XGB). Llama2-13B scored an AUROC of 58.54 on CHF prediction, the best among all models. This demonstrates that LLM embeddings can match or nearly match the performance of models trained with raw data inputs.

\subsection{Main results for mortality prediction and length-of-stay}

\begin{figure}[h!]
    \centering
    \includegraphics[width=\columnwidth]{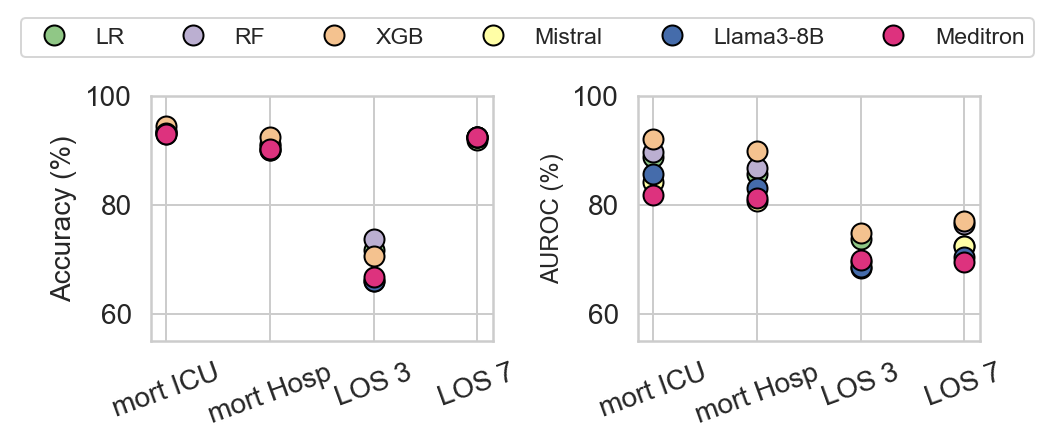}
    \vspace{-.3in}
    \caption{\small Accuracy (left) and AUROC (right) for in-ICU mortality (mort ICU), in-Hospital morality (mort Hosp), hospital LOS exceeding 3 days (LOS 3) and 7 days (LOS 7). The Logistic Regression (LR) and Random Forest (RF) baselines are reported from~\cite{wang2020mimic}. The LLM results are from LLM embeddings + XGB settings. The CIs mostly overlap; for clarity in presentation, they were omitted from this figure. }
    \label{fig:mort_los}
\end{figure}

Figure~\ref{fig:mort_los} displays performance for various models on tasks of in-ICU mortality (mort ICU), in-hospital mortality (mort Hosp), and hospital length-of-stay for more than 3 (LOS 3) and 7 days (LOS 7). The raw data features with XGB model consistently outperforms others with an AUROC of 92.02 in mort ICU and 89.83 in mort Hosp. 
LLM embeddings from Mistral, Llama3-8b, and Meditron, while slightly lagging behind the raw data features with ML classifiers in the mortality tasks, performed comparably in the LOS 7 tasks. Mistral with XGB achieved accuracy of 92.34 and AUROC of 72.36 on LOS 7 task, showing competitive performance to XGB with raw data features, with accuracy of 92.32 and an AUROC of 76.93. 
The gap between LLM embeddings and raw data for mortality and LOS tasks suggests a need to improve time-varying feature representation. 

%\subsection{Comparison across different embedding extraction and format conversion methods} 

\subsection{Comparisons across different embedding methods and data conversion methods}

\begin{figure}[h!]
    \centering
    \includegraphics[scale=0.43]{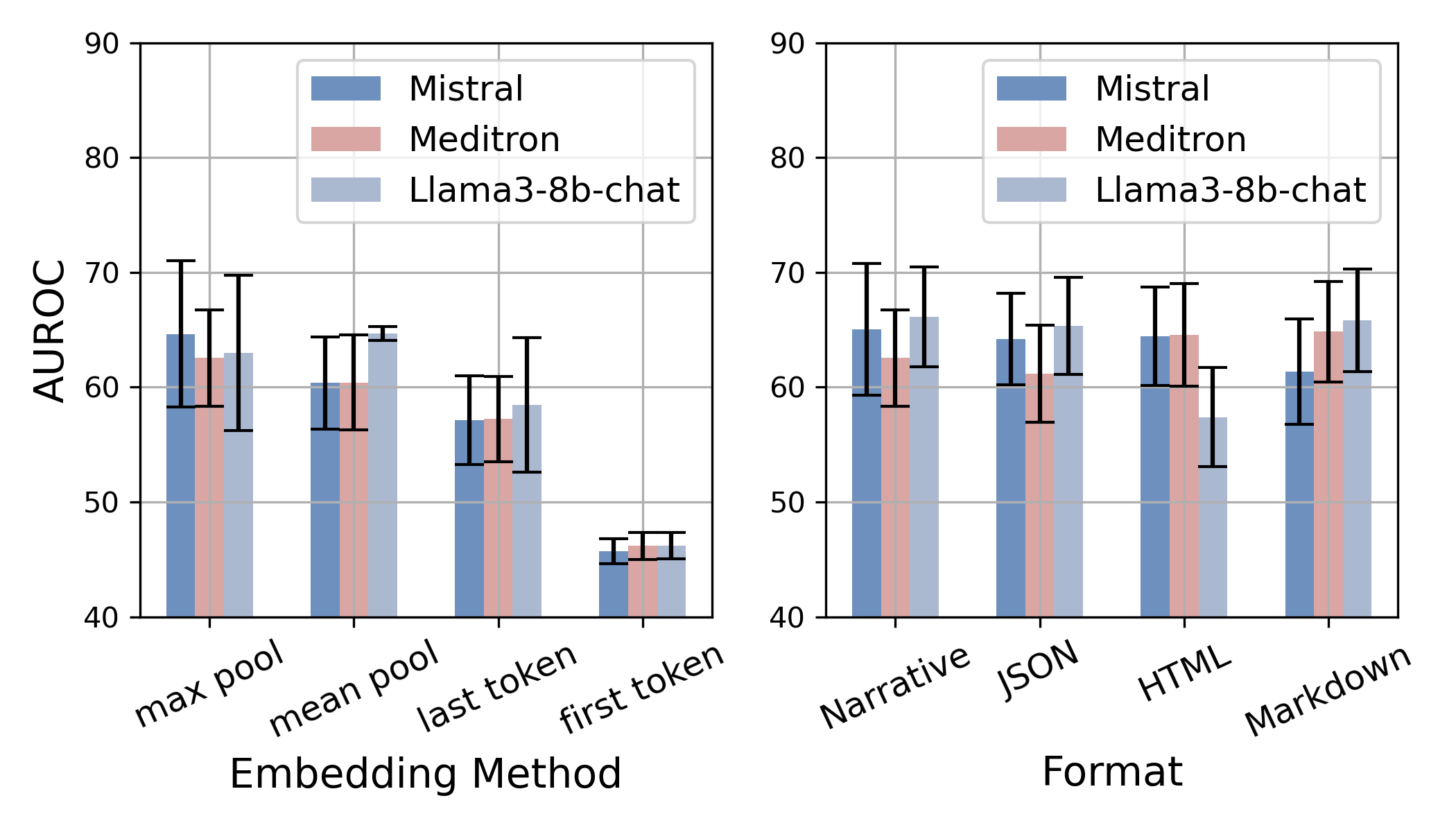}
    \vspace{-.26in}
    \caption{\small Comparison across different embedding methods and different format on the Diagnosis dataset. For simplicity, we used \textsc{Narrative} and max pooling for the other analysis after this section.  }
    \label{fig:embedding_format}
\end{figure}

Figure~\ref{fig:embedding_format} presents AUROC values for different embedding methods and data conversion formats across three models: Mistral, Meditron, and Llama3-8b. Max pooling achieves the highest performance for Mistral (64.62) and Meditron (62.54), while mean pooling is most effective for Llama3-8b (64.69). The last token method yields moderate performance across all models, with AUROCs around 57, while first token embeddings result in the lowest AUROC values, indicating a less effective representation for these models. 

When encoding data with different formats, Mistral shows preference for \textsc{Narratives, Json}, and \textsc{Html}. The \textsc{Markdown} format generally yielded the lowest performance across the models, particularly for Mistral. \textsc{Json} and \textsc{Html} formats showed competitive performance, with JSON being slightly more effective for Meditron and Llama3-8b. Notably, Llama3-8b exhibited the highest variability across formats, with AUROCs ranging from 57.40 (\textsc{Html}) to 66.13 (\textsc{Narratives}). %\footnote{For simplicity, we used \textsc{Narrative} and max pooling for other analysis. }

% \begin{table}[h]
% \small 
% \centering
% \begin{tabular}{lccc}
% \toprule
% Method & Mistral & Meditron & Llama3-8b-chat \\
% \midrule
% max pool & 64.62 ± 6.37 & 62.54 ± 4.21 & 62.97 ± 6.76 \\
% mean pool      & 60.37 ± 4.03 & 60.42  ± 4.12 & 64.69 ± 0.60 \\
% last token      & 57.15 ± 3.87 & 57.24 ± 3.72  & 58.47 ± 5.85\\
% first token  & 45.71 ± 1.09 & 46.18 ± 1.16  & 46.20 ± 1.15   \\
% \bottomrule
% \end{tabular}
% \label{tab:embedding_auroc}
% \vspace{-.1in}
% \caption{\small Comparison of AUROC values for different embedding methods and models on Diagnosis dataset using \textsc{narrative} data format.}
% \end{table} 

% \begin{table}[h]
% \small 
% \centering
% \begin{tabular}{lccc}
% \toprule
% Format & Mistral & Meditron & Llama3-8b-chat \\
% \midrule
% Narrative & 65.04 ± 5.75 & 62.54 ± 4.21 & 66.13 ± 4.33 \\
% JSON      & 64.19 ± 3.97 & 61.18 ± 4.24 & 65.33 ± 4.24 \\
% HTML      & 64.45 ± 4.29 & 64.54 ± 4.46 & 57.40 ± 4.33 \\
% Markdown  & 61.35 ± 4.59 & 64.84 ± 4.37 & 65.81 ± 4.48 \\
% \bottomrule
% \end{tabular}
% \label{tab:format_auroc}
% \vspace{-.1in}
% \caption{\small Comparison of AUROC values for different formats and models, using max pooling embedding method.}
% \end{table}

\subsection{Impact of prompt engineering and few-shot learning} 
%\subsection{Effect of prompt engineering and few-shot learning}  

% \begin{figure}
%     \centering
%     \includegraphics[scale=0.34]{latex/Figs/auroc_sys_cot_fs.png}
%     \vspace{-.35in}
%     \caption{\small The impact of system prompts, Chain-of-thought and pairs of positive and negative examples as few-shot on the LLM performance on Diagnosis dataset. \textcolor{red}{need to add which sys prompt was pictured.}} 
%     \label{fig:promt_cot_fewshot}
% \end{figure}

% \begin{table}
% \small 
% \centering
%     \begin{tabular}{c|ccc}
%        Model  & Setting & Diagnosis & AUROC  \\
%          & 
%     \end{tabular}
% \end{table}

We compared performance of Mistral and Llama3 using different system instructions under zero-shot and few-shot settings, as well as CoT examples. Mistral, under 0-shot with a system instruction with persona of medical professional and the task of assessing patient condition (prompt 1 in Table~\ref{tab:prompts}), achieved an AUROC of 71.35 on Sepsis prediction, the highest of all models. Llama3 with zero-shot prompting using prompt 1 in Table~\ref{tab:prompts} showed reported AUROC of 73.51 on Arrhythmia, surpassing its counterpart at 71.08 but still below raw data XGB baseline (76.49). CoT and few-shot exhibited various performance and often resulted in lower AUROC scores compared to Table~\ref{tab:ml_baselines_sepsis_arrhythmia_chf_prediction}. Full results are provided in Appendix~\ref{sec:prompt_design}.    
%Figure~\ref{fig:promt_cot_fewshot} compares the performance of Mistral and Llama3 models using different techniques (zero-shot with system prompts and few-shot with and without CoT examples) across three tasks on Diagnosis dataset. The main findings indicate that using zero-shot LLMs can achieve better results than few-shot settings. Specifically, Mistral with 0-shot prompting outperformed the best overall performance for Sepsis, achieving an AUROC of 71.35 compared to the raw data XGB baseline's 71.17. Llama3 with zero-shot prompting showed strong results for Arrhythmia, surpassing its non-prompted version. In contrast, CoT and few-shot techniques exhibited variable performance but did not surpass the system prompts, often resulting in lower AUROC scores. In conclusion, LLM embeddings are sensitive to the system prompts and few-shot examples. %carefully designed system prompts significantly enhance model performance, underscoring the importance of prompt engineering for improving LLM embeddings in diagnostic tasks. 

\subsection{Parameter efficient fine-tuning results} 

\begin{table}[h!]
\scriptsize 
    \centering
    \begin{tabular}{c|ccccc} \toprule
      Setting  & LOS 3 & LOS 7 & Mort ICU & Mort Hosp \\ \midrule 
      Mistral   & 67.84 & 72.36 & 84.16 & 80.71  \\ 
      Mistral$_{QLoRA}$ & 65.26 & 67.66 & 75.69 & 73.66  \\ 
     Performance Drop & 2.58 & 4.70 & 8.47 & 7.05  \\ \midrule
     Llama3-8b & 68.54& 70.38 & 85.61 & 83.06 \\
     Llama3-8b$_{QLoRA}$ & 66.69 & 68.56 & 75.14 & 71.15 \\ 
     Performance Drop & 1.85 & 1.82 & 10.47 & 11.91 \\ 
      \bottomrule
    \end{tabular}
    \vspace{-.1in}
    \caption{\small AUROC comparison before and after training LLM with QLoRA on MIMIC tasks.}
    \label{tab:qlora_mistral}
\end{table}

\begin{figure}
    \centering
    \includegraphics[scale=0.28]{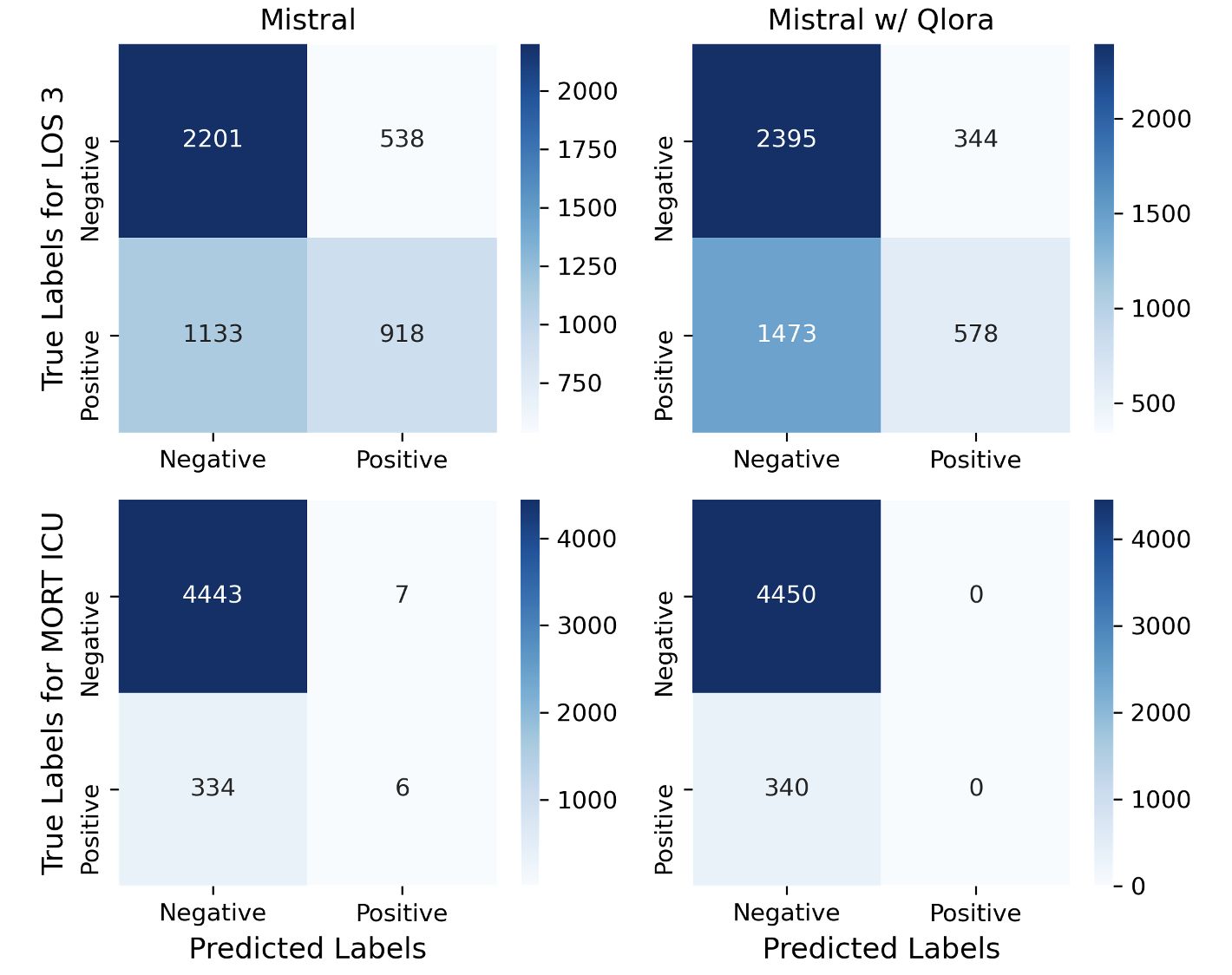}
     \vspace{-.1in}
    \caption{\small Confusion matrices for Mistral prediction on LOS 3 and Mort ICU tasks. Right: Mistral without QLoRA; left: Mistral after QLoRA.}
    \label{fig:conf_mat_qlora_mistral}
\end{figure}

Table~\ref{tab:qlora_mistral} presents results of Mistral and Llama3-8b under the QLoRA across all four tasks from MIMIC-Extract. The performance drops are noticeable, especially in the two mortality predictions. To further understand the reason behind the performance drops, we plotted the confusion matrices for LOS 3 and Mort ICU, comparing Mistral's predictions before and after QLoRA in Figure~\ref{fig:conf_mat_qlora_mistral}. For LOS 3 prediction, the Mistral model with QLoRA shows an increase in true negatives and a decrease in false positives. However, the false negatives rises from 1133 to 1473, and true positive drops from 918 to 578. On the Mort ICU task, the Mistral model with QLoRA correctly predicts no false positives, but it fails to predict any positive cases (0 true positives). The performance drop can be attributed to the imbalanced class distribution in the dataset, as the models show a tendency to favor the majority class (negative cases). During QLoRA, the LLM might learn the class prevalence, biasing its representation and making it challenging to correctly identify the minority class. %This imbalance makes it challenging for the model fine-tuned with QLoRA to correctly identify the minority class, thereby impacting overall performance. 

\subsection{LLM Embedding vs LLM Generation} 

The final experiments compare the performance of LLM embeddings combined with ML classifiers against direct outputs from LLMs. This comparison shows that, although LLM embeddings generally do not outperform raw data features, they offer a more robust and reliable solution than relying on LLMs to directly answer Yes or No questions. Our exploration revealed significant limitations in LLM generation for binary prediction tasks. For instance, Mistral frequently predicted 'Yes' for sepsis, arrhythmia, and CHF AORC, resulting in AUROC scores being 50, whereas LLM embeddings achieved AUROCs of 71.12 for sepsis, 72.26 for arrhythmia, and 63.54 for CHF AUROC. Similar patterns were observed from Llama3-8b results (Table~\ref{tab:llm_gen_diagnosis}). On MIMIC-Extract tasks with highly skewed class distributions, Mistral and Llama3-8b, when generating direct Yes/No answers, again showed reduced ability to discriminate between positive and negative cases (Table~\ref{tab:llm_gen_mimic}). These findings underscore the need for embeddings, which provide a more nuanced and effective approach for clinical predictions. We refer readers to Appendix~\ref{sec:llm_gen} for more details.

%\input{latex/rq2_influence}
%\input{latex/rq3_fewshot}
%\input{latex/new_methods}
%\input{latex/embedding}

\section{Discussion}
\begin{figure}
    \centering
    \includegraphics[scale=0.4]{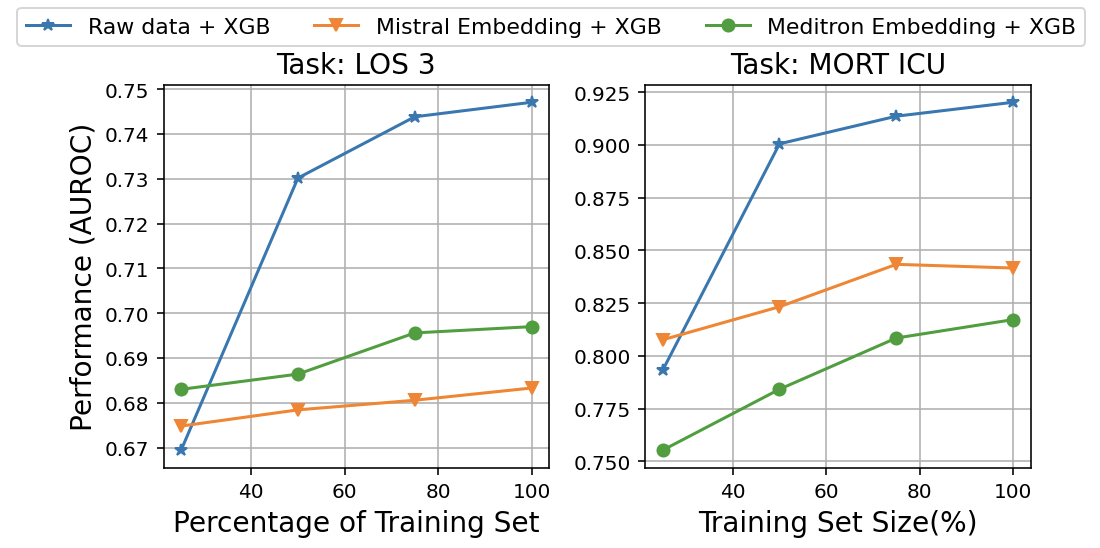}
    \vspace{-.20in}
    \caption{\small AUROC comparison between Raw data, embeddings from Mistral and Meditron with XGB classifiers, by controlling the training set size on two MIMIC tasks.}
    \label{fig:training_sensitivity}
\end{figure}

%\textbf{Sensitivity to training set size} 
To understand the discrepancy between the two data representations, we examined the training effectiveness of raw data features and LLM embeddings by controlling the training set size. 
Figure~\ref{fig:training_sensitivity} compares the performance of the raw data XGB baseline model with the Mistral and Meditron embeddings across different training set sizes for two tasks in the MIMIC dataset. The raw data XGB baseline model shows a significant increase in AUROC scores with larger training sets, achieving high performance. In contrast, both the Mistral and Meditron embeddings paired with XGB models exhibit much smaller improvements, consistently performing lower than the raw data XGB baseline. This highlights the greater effectiveness of XGB when learning from raw data features compared to LLM embeddings for these prediction tasks.

Our findings suggest that raw data features provide more informative input for ML models compared to LLM-generated embeddings. While LLM embeddings capture complex representations, they may not be as tailored for binary medical prediction tasks. Additionally, computing efficiency is an important consideration, as LLMs require significantly more GPU memory than raw data features.

However, zero-shot LLM embeddings achieve comparable performance in certain scenarios, highlighting their potential for rapid deployment without extensive training. A promising direction is distilling these embeddings into a smaller space while retaining their extensive knowledge~\cite{lee2024gecko}. \citeauthor{behnamghader2024llm2vec}~(\citeyear{behnamghader2024llm2vec}) recently proposes LLM2Vec, a method to train decoder-only LLMs as text encoders with unsupervised training, which merits further investigation. 
\section{Conclusion}

We present the first analysis of LLM embeddings for numerical EHR data features in medical ML applications, showing the opportunity and challenges of using LLM embeddings as a substitute of raw data features. %Although raw data remains dominant, LLM embeddings show substantial potential. 
We hope to encourage future research on improving LLM embeddings, particularly for imbalanced label prediction, and advancing health predictions with multi-modal data, while addressing interpretability and bias. 
\newpage 

\section{Limitation}
In our study, we focused on investigating some of the most common LLMs, including Meditron, Mistral, Llama2, and Llama3. Due to GPU constraints, some experiments, such as Qlora, were conducted on only one or two models, limiting the comprehensiveness of our analysis. We did not include black-box LLMs via API because, despite using fully de-identified data, both EHR datasets are protected under Data Use Agreement, restricting us sharing with third parties. Additionally, we acknowledge that we did not explore all possible methods of prompting LLMs, which may have influenced our results. Furthermore, our examination was restricted to the last layers of the LLMs, potentially overlooking valuable information encoded in other layers.

Regardless of these limitations, our findings are consistent across models: zero-shot LLM embeddings paired with machine learning classifiers generally underperform compared to raw data features, though they sometimes achieve comparable performance. 

% A key limitation is our study's focus on comparing LLMs solely with XGBoost for EHR tabular data analysis, overlooking other prevalent ML algorithms like SVMs, Random Forests, or Logistic Regression. This may affect the generalizability of our findings within the vast landscape of machine learning in medical diagnostics. Furthermore, focusing on XGB and LLM probability correlations may narrow our insight, omitting a richer evaluation of model performance and interpretability across diverse metrics. 

\section{Ethical Statement}

Following the ACL's ethical review guidelines, our study on leveraging LLMs for medical diagnosis within EHR emphasizes ethical integrity by prioritizing harm avoidance, privacy protection, fairness, transparency, and respect for intellectual property. While our research aims to advance medical diagnostics through LLMs, there is a potential risk that misinterpretations of model predictions could inadvertently lead to diagnostic errors or bias in clinical decision-making. Therefore, rigorous validation protocols, including expert medical review and bias detection mechanisms are needed to ensure that model predictions are both accurate and equitable across diverse patient populations. 

We have rigorously ensured data de-identification, obtained ethical approvals, actively mitigated biases, and maintained openness in our methodologies and findings to uphold honesty and reproducibility. Our commitment extends to respecting intellectual property through proper attribution and license adherence, with the overarching goal of contributing positively to healthcare outcomes and societal well-being. This approach underscores the importance of robust, secure research practices in developing computational tools for healthcare, aligning with our ethical responsibility to advance the field for the public good.

\bibliography{custom}
\bibstyle{acl_natbib}

\appendix

\section{Diagnosis Prediction Dataset Patient Demographics} 
\label{sec:patient}

\begin{table}[ht]
\small 
    \centering
    \begin{tabular}{l|l} \toprule
       Group  &  Distribution \\ \midrule 
       Total & 660 Patients \\ 
     Gender & Male (52\%), Female (48\%) \\ 
     Ages & Adults (36\%), Geriatric (64\%)\\
     Race & White\textbackslash Caucasian (89\%), Black\textbackslash African\\
     & American (6\%), Asian\textbackslash Mideast Indian  \\
     & (2\%), American Indian\textbackslash Alaska Native \\ 
     & (1\%), Pacific Islander\textbackslash Hawaiian Native\\
     &  (<1\%) , Declined\textbackslash Unknown (<1\%) \\ 
    % Clinical features & \\ 
    \midrule 
    \end{tabular}
    \vspace{-.14in}
    \caption{\small Patient demographic description of diagnosis prediction dataset. Note that in this work, we exclude the demographic information from ML input. }
    \label{tab:stat_diag}
\end{table}

\section{Probing LLMs for Inherent Knowledge of Normal Ranges} 
\label{sec:prob_examples}

A foundational question for using LLM embeddings for numerical data representation is whether they possess inherent knowledge about the normal range of values for clinical data. To assess this, we first asked the LLMs about standard physiological ranges, measurement units, and reasoning from the feature set of tabular data outlined in Table~\ref{tab:feat_and_templates}. A board-certified physician assessed the LLM generations using a 1 to 5 Likert scale across three dimensions: correctness of the range, accuracy of measurement units, and quality of explanations ("Reasoning"). Our probing experiments were conducted on Mistral and Llama2-13b, as these are general-domain LLMs that have been fine-tuned to follow instructions. This experiment was done prior to the release of Llama3 models, therefore we did not include them. We did not employ automated metrics because there is no single set of reference ranges for some features, as they are highly dependent on patients' ages, sex, and other demographic factors.  

In our exploration of language model capabilities within the healthcare domain, we specifically probe the model's understanding of standard ranges and measurement units for EHR data features. The following table illustrates Mistral's performance on tasks designed to evaluate its ability to accurately identify and articulate the normal physiological ranges and corresponding units for 24 distinct EHR features. This assessment not only tests the model's grasp of medical terminology and concepts but also its potential application in clinical settings where precise knowledge of such parameters is crucial.

\begin{table}
\small
\centering
\begin{tabularx}{\columnwidth}{X} 
\toprule 
\textbf{Question:} What is the reference range for the Pulse Pressure Index? \\
\textbf{Reference:} 0.1, 1 \\
\textbf{LLM (Mistral):} The reference range for Pulse Pressure Index (PPI) is not a commonly used vital sign or lab result label. It measures the ratio of the pulse pressure to the systolic blood pressure, used to assess cardiovascular risk. The PPI is typically reported as a percentage, with a normal range being less than 10\%. However, the specific reference range may vary by population and clinical context. It is important to consult healthcare professionals for personalized PPI interpretations. \\
\textbf{Physician Evaluation (Range, Unit, Reasoning):} 3, 5, 5 \\
\midrule 
\textbf{Question:} What is the reference range for Serum Carbon Dioxide Level? \\
\textbf{Reference:} 22, 29 mmol/L \\
\textbf{LLM (Mistral):} The reference range for serum carbon dioxide (CO2) level is typically between 21 to 29 millimoles per liter (mmol/L) for serum measurements, or 2.5 to 3.5 mmol/L for arterial blood gas (ABG) measurements. Note that CO2 level units are mmol/L or ppm, and the range may vary by laboratory or tested population. \\
\textbf{Physician Evaluation (Range, Unit, Reasoning):} 5, 4, 2 \\ 
\bottomrule 
\end{tabularx}
\caption{Mistral's probing output examples for Normal Ranges and Units of EHR data features}
\label{tab:normal_range}
\end{table}

\section{Prompt Design}
\label{sec:prompt_design}

\begin{table*}[ht!]
\small 
\centering
\begin{tabularx}{\textwidth}{@{}X@{}}
\toprule
 \textbf{Prompt Description} \\
\midrule
Prompt 1 -\textbf{Persona: Medical Professional} As a healthcare provider, please assess the patient's condition provided below and outline the likely causes or diagnoses for their clinical worsening. List only the diagnoses and keep your response brief. \\
\addlinespace[0.25em]
Prompt 2 -\textbf{Persona: AI System} You are an AI with medical expertise. Create an embedding for the probable problems or diagnoses that are causing clinical deterioration, based on the patient's condition detailed below, to aid in training a diagnostic prediction machine learning classifier. Be brief in your description. \\
\addlinespace[0.25em]
Prompt 3 -\textbf{Persona: Medical Professional (Chain-of-Thought)} As a medical expert, please examine the patient's condition by first identifying any abnormal values. Next, critically analyze these values to assess their impact, and clearly state your final diagnosis regarding what might be causing the clinical deterioration. Keep your summary brief. \\
\addlinespace[0.25em]
Prompt 4 -\textbf{Persona: Medical Professional (Binary Question)} You are a medical doctor. Based on the patient’s condition, determine the likelihood that diagnosis X is causing their clinical deterioration. Be aware that diagnosis X occurs in Y\% of similar cases. \\ 
\bottomrule
\end{tabularx}
\vspace{-.1in}
\caption{\small System prompts for medical diagnosis assistance with different persona settings.}
\label{tab:prompts}
\end{table*} 

To test the impact of prompt engineering on LLM embeddings, we designed four distinct prompts, varying by persona, thinking style, and question types. Our prompt engineering strategy builds on prior work that utilizes perplexity to select optimal prompts~\cite{gonen2023demystifying}. Initially, we crafted meta-language prompts delineating clinical scenarios and tasks. We then employed GPT-4, a state-of-the-art LLM, to generate five paraphrases for each prompt to capture a range of expressions. We evaluated each paraphrase's naturalness and fluency by calculating its perplexity score, selecting the most coherent paraphrase for each scenario for our experiments. The optimized prompts are detailed in Table~\ref{tab:prompts}.

We also used these prompts in few-shot experiments. Given the complexity of the MIMIC-Extract dataset, which includes 104 features with data collected over 24 hours, creating high-quality synthetic data is challenging. Thus, we focused our efforts on the diagnosis prediction use case, which involves simpler inputs of twenty-four clinical features, %making quality control possible. 
facilitating quality control over the synthetic data generation. %more reliable synthetic data quality assessments. 

As illustrated in Figure~\ref{fig:study}, we used GPT-4 to create synthetic data depicting patient cases of clinical deterioration with features in Table~\ref{tab:feat_and_templates}. This data set includes both positive and negative diagnosis cases, which were reviewed by an expert physician and clinical informaticist for quality assurance. Our few-shot experiments varied in complexity: the first modified Prompt 1 to include example input-output pairs ("Simple" few-shot setting), while the second added a CoT explanation detailing the diagnostic reasoning into Prompt 3. The CoT was structured to identify and reason over abnormal values to conclude diagnoses, enhancing the data's interpretability and educational value. 

\begin{table}[h]
\centering
\small 
\begin{tabular}{lcc}
\toprule
\textbf{Model} & \textbf{Setting} & \textbf{AUROC CI (\%)} \\ 
\midrule
Mistral-7b-instruct & sys1 &  54.85 [48.18, 62.11] \\ 
 & sys2 &  53.88 [47.37, 60.67] \\ 
 & sys3 &  51.16 [44.34, 57.67] \\ 
 & sys4 &  54.04 [46.84, 61.56] \\ 
 & Fewshot &  54.43 [46.62, 61.05] \\  
& CoT & 57.96 [60.72, 69.24] \\ \midrule
Llama2-13b-chat & sys1 &  56.49 [49.90, 63.09] \\ 
 & sys2 &  55.61 [48.43, 62.31] \\ 
 & sys3 &  50.41 [43.19, 57.33] \\ 
 & sys4 &  60.24 [53.28, 67.09] \\ 
 & Fewshot & 53.12 [46.38, 59.84] \\ 
 & CoT & 54.10 [51.84, 60.59]\\ \midrule
Llama3-8b-instruct & sys1 &  52.81 [46.47, 59.21] \\ 
 & sys2 &  51.11 [44.64, 57.27] \\ 
 & sys3 &  49.03 [42.19, 55.74] \\ 
 & sys4 & 55.23 [48.28, 61.79] \\ 
 & Fewshot &  53.24 [46.60, 59.99] \\ 
 & CoT & 51.44 [48.58, 53.77] \\ 
\bottomrule
\end{tabular}
\vspace{-.1in}
\caption{\small AUROCs for various models and settings on CHF Volume Overload prediction.}
\label{tab:auroc_chf_volumeoverload}
\end{table}

\begin{table}[h]
\centering
\small  
\begin{tabular}{lccc}
\toprule
\textbf{Model} & \textbf{Setting} & \textbf{AUROC CI (\%)} \\ 
\midrule
Mistral-7b-instruct & sys1 &  62.27 [56.47, 67.67] \\ 
 & sys2 &  63.84 [58.06, 69.31] \\ 
 & sys3 &  64.92 [58.98, 70.05] \\ 
 & sys4 &  66.11 [60.20, 71.52] \\ 
 & Fewshot &  68.43 [62.82, 73.98] \\ 
\midrule

Llama2-13b-chat & sys1 & 69.24 [63.52, 74.84] \\ 
 & sys2 &  61.90 [56.04, 67.63] \\ 
 & sys3 &  61.44 [56.35, 66.68] \\ 
 & sys4 &  64.43 [58.63, 69.95] \\ 
 & sys5 &  67.74 [62.35, 73.22] \\ 
\midrule
Llama3-8b-instruct & sys1 & 71.12 [65.91, 76.05] \\ 
 & sys2 & 72.13 [66.12, 77.88] \\ 
 & sys3 & 70.24 [64.94, 75.58] \\ 
 & sys4 & \textbf{73.51 [68.09, 78.54]} \\ 
 & sys5 &73.10 [67.29, 78.18] \\ 
\bottomrule
\end{tabular}
\vspace{-.1in}
\caption{\small One Time AUROC and Confidence Intervals for various models and settings on Arrhythmia prediction. Scores are multiplied by 100.}
\label{tab:auroc_arrhythmia}
\end{table}

\begin{table}[h]
\centering
\small 
\begin{tabular}{lccc}
\toprule
\textbf{Model} & \textbf{Setting} & \textbf{AUROC CI (\%)} \\ \midrule
Mistral-7b-instruct & sys1 & \textbf{71.35 [67.39, 75.73]} \\ 
 & sys2 & 67.63 [63.53, 71.80] \\ 
 & sys3 & 65.67 [61.50, 69.78] \\ 
 & sys4 & 67.87 [63.29, 71.88] \\ 
 & Fewshot & 67.32 [63.26, 71.20] \\ 
   & CoT & 64.29 [60.72, 69.24] \\ \midrule
Llama2-13b-chat & sys1 & 68.79 [64.87, 72.59] \\ 
& sys2 & 69.82 [65.66, 73.92] \\ 
 & sys3 & 68.92 [64.64, 73.09] \\
& sys4 & 64.62 [60.40, 68.73] \\ 
 & Fewshot & 66.49 [62.21, 70.59] \\ 
  & CoT & 65.13 [62.30, 69.25] \\ \midrule
Llama3-8b-instruct & sys1 & 67.05 [62.96, 71.11] \\ 
 & sys2 & 66.07 [61.94, 70.28] \\ 
 & sys3 & 64.80 [60.74, 69.11] \\ 
 & sys4 & 66.81 [63.08, 70.90] \\ 
 & Fewshot & 66.87 [62.90, 70.84] \\ 
 & CoT & 62.12 [58.37, 66.96] \\ 
\bottomrule
\end{tabular}
\vspace{-.1in}
\caption{\small AUROC Confidence Intervals for various models and settings on Sepsis prediction. Scores are multiplied by 100.}
\label{tab:auroc_sepsis}
\end{table}

% \begin{table}[h!]
%     \centering
%     \small
%     \begin{tabular}{c|ccc} \toprule 
%        Model  & Sys Prompt & Diagnosis & AUROC  \\
%        Best  &  - & Sepsis & 71.17 \\ 
%        Mistral & Sys 1 & & \textbf{72.11} \\ 
%        Mistral & CoT & & 64.29 \\ 
%        Mistral & FewShot& & 67.32 \\ 
%        Llama3 & Sys 2 & & 69.83 \\ 
%        Llama3 & CoT & & 62.12 \\ 
%        & FewShot & & 66.87 \\ \midrule 
%        Best  &  - & Arrhythmia & 76.49 \\ 
%        Mistral & Sys4 & & 66.11 \\ 
%         Mistral & CoT & & 64.29 \\ 
%        Mistral & FewShot& & 68.43 \\
%        Llama3 & Sys 2 &  & 72.13 \\ 
%        Llama3 & CoT & & 63.96 \\ 
%        & FewShot & & 67.74 \\\midrule 
%        Best & - & CHF & 63.84 \\ 
%        Mistral & Sys1 & & 54.80 \\ 
%        Mistral & CoT & & 57.96 \\
%         Mistral & FewShot & & 54.43 \\
%        Llama3 & Sys4 & &60.24 \\ 
%        Llama3 & CoT & &51.44 \\
%         Llama3 & FewShot & & 53.12 \\
%        \bottomrule 
%     \end{tabular}
%     \caption{Caption}
%     \label{tab:my_label}
% \end{table}

\begin{table}[ht!]
\small 
    \centering
    \begin{tabular}{cccc} \toprule 
        Model &  Task & AUROC & Accuracy\\ \midrule
        Mistral-7b-instruct & Sepsis & 50.00 & 43.18 \\ 
        & Arrythmia & 50.00 & 15.30 \\ 
        & CHF & 50.00 & 11.82 \\ \midrule
        Llama3-8b-instruct & Sepsis & 50.28 & 54.69 \\ 
        & Arrythmia & 47.12 & 73.63 \\
        & CHF & 47.61 & 77.12 \\ 
        \bottomrule
    \end{tabular}
    \caption{Results of Mistral and Llama3-8B directly generating "Yes/No" to the Diagnosis dataset. To align with the results reported for emebdding+ML classifiers settings, the LLMs are zero-shot, and no additional system instructions are included in the chat template. }
    \label{tab:llm_gen_diagnosis}
\end{table}

\begin{table}[ht!]
\small 
    \centering
    \begin{tabular}{cccc} \toprule
        Model &  Task & AUROC & Accuracy\\ \midrule
        Mistral-7b-Inst & LOS 3 & 49.72 & 56.05 \\ 
        & LOS 7 & 49.86 & 87.93 \\ 
        & MORT ICU & 50.04 & 91.98 \\
        & MORT HOSP & 49.79 & 86.93 \\
        % Meditron & LOS 3 & 49.84 & 42.89 \\ 
        % & LOS 7 & 49.84 & 7.83 \\ 
        % & MORT ICU & 50.00 & 7.12   \\
        % & MORT HOSP & 50.00 & 10.27 \\
        Llama3-8b-Inst & LOS 3 & 50.73 & 57.24 \\ 
        & LOS 7 & 50.10 & 92.29 \\
        & Mort ICU & 49.99 & 92.88 \\
        & Mort Hosp & 49.99 & 89.71 \\ 
        \bottomrule
    \end{tabular}
    \caption{Results of Mistral and Meditron direct generation of "Yes/No" to the MIMIC-Extract dataset}
    \label{tab:llm_gen_mimic}
\end{table}

\section{Results of LLM Direct Generation}
\label{sec:llm_gen}
We tested the ability of Mistral and Llama3 to directly predict Yes or No answers to questions from the Diagnosis and MIMIC-Extract datasets. To achieve this, we added specific instructions directing the LLMs to respond only with "Yes" or "No," then parsed the outputs to 1 or 0 labels and computed AUROC and Accuracy. For this experiment, we set the maximum token limit to 25 and the top k to 50. 

Table~\ref{tab:llm_gen_diagnosis} presents results of Mistral directly generating "Yes/No" answers for the Diagnosis dataset. For all tasks (Sepsis, Arrhythmia, CHF), Mistral achieved an AUROC of 50.00, indicating no discriminatory ability. Accuracy varied across tasks, with Sepsis at 43.18\%, Arrhythmia at 15.30\%, and CHF at 11.82\%, corresponding to the positive class distribution, demonstrating poor performance in direct prediction. Llama3 exihibited similar performance: it reported AUROC scores between 47.12 (Arrhythmia) to 50.28 (Sepsis), underperforming its embedding counterparts reported in Table~\ref{tab:ml_baselines_sepsis_arrhythmia_chf_prediction}.
Admittedly, extra effort in prompt engineering and parameter searching could improve direct generation results. However, compared to their embedding + ML classifier counterparts in the same zero-shot setting and input format (\textsc{Narratives}, without additional system instructions such as personas), their performance is significantly lower. 

On the MIMIC-Extract tasks, table~\ref{tab:llm_gen_mimic} shows the results of Mistral-7b-Instruct and Llama3-8b-Instruct in directly generating "Yes/No" answers for various tasks in the MIMIC-Extract dataset. Both models demonstrated no discriminatory ability, with AUROC scores close to 50 for all tasks. Accuracy varied, with notable high accuracy for MORT ICU and MORT HOSP tasks, particularly for Llama3-8b-Instruct (92.88\% and 89.71\%, respectively). However, these high accuracy scores likely reflect class imbalance rather than model performance. The contrast between the LLM direct prediction performance and LLM embedding + classifier performance further suggests that LLM embeddings provide a more robust method.

\section{Parameter Grids for ML Classifiers}
\label{sec:param_grid}

\begin{table}[h]
\centering
\begin{tabular}{|l|l|}
\hline
\textbf{Parameter}      & \textbf{Values}             \\ \hline
\texttt{n\_estimators}  & 50, 100, 250, 500           \\ \hline
\texttt{max\_depth}     & 2, 5, 10, 15, 20            \\ \hline
\texttt{learning\_rate} & 0.005, 0.01, 0.05, 0.1      \\ \hline
\texttt{min\_child\_weight} & 1, 2, 3               \\ \hline
\end{tabular}
\caption{Parameter grid for XGBoost (XGB) classifier.}
\label{tab:param_grid}
\end{table}
 
\begin{table}[h]
\centering
\begin{tabular}{|l|l|}
\hline
\textbf{Parameter} & \textbf{Values}     \\ \hline
\texttt{alpha}     & 0.1, 0.5, 1.0       \\ \hline
\texttt{l1\_ratio} & 0.1, 0.5, 0.9       \\ \hline
\end{tabular}
\caption{Parameter grid for Logistic Regression (LR).}
\label{tab:param_grid_logistic_regression}
\end{table}

We conducted a comprehensive grid search for hyperparameter optimization on two classifiers: XGBoost (XGB) and Logistic Regression. For the XGB classifier, the parameter grid included $n estimators$ set to [50, 100, 250, 500], $max depth$ ranging from [2, 5, 10, 15, 20], $learning rate$ values of [0.005, 0.01, 0.05, 0.1], and $min child weight$ values of [1, 2, 3]. This extensive search aimed to identify the best combination of hyperparameters to enhance model performance.

For the Logistic Regression classifier, we varied alpha with values of [0.1, 0.5, 1.0] and $l1 ratio$ with values of [0.1, 0.5, 0.9]. This grid search was designed to fine-tune the regularization parameters to achieve optimal balance between model complexity and performance.

Grid-searching on XGB parameters took 25-40 minutes on GPU. On LR, it took about 25 minutes to search for the best parameters. Training both classifiers took less than 5 minutes, even on the MIMIC-Extract dataset where there are more than 16000 samples.

\label{sec:appendix}

\end{document}